\documentclass{article}
\usepackage{ijcai20}

\usepackage{times}
\usepackage{soul}
\usepackage{url}
\usepackage[hidelinks]{hyperref}
\usepackage[utf8]{inputenc}
\usepackage[small]{caption}
\usepackage{graphicx}
\usepackage{amsmath}
\usepackage{amsthm}
\usepackage{booktabs}
\usepackage[shortlabels]{enumitem}
\usepackage[TABBOTCAP]{subfigure}
\usepackage{multirow}
\usepackage{algorithm}
\usepackage{algpseudocode}
\usepackage[normalem]{ulem}
\usepackage[colorinlistoftodos]{todonotes}
\usepackage{comment}

\usepackage{color}

\def\CM{{\mathcal C}}

\def\GM{{\mathcal G}}

\def\KM{{\mathcal K}}

\def\PM{{\mathcal P}}

\def\RM{{\mathcal R}}

\def\e{{\bf e}}

\def\o{{\bf o}}

\def\0{{\bf 0}}
\def\1{{\bf 1}}

\newtheorem{definition}{Definition}

\title{
    TransOMCS: From Linguistic Graphs to Commonsense Knowledge
}

\author{Hongming Zhang$^1$, Daniel Khashabi$^2$, Yangqiu Song$^1$, Dan Roth$^3$\\
\affiliations
$^1$The Hong Kong University of Science and Technology,\\  $^2$Allen Institute for AI, $\quad$ $^3$University of Pennsylvania\\  
\emails
hzhangal@cse.ust.hk, $\quad$ danielk@allenai.org, $\quad$yqsong@cse.ust.hk, $\quad$ danroth@seas.upenn.edu
}

\date{}

\begin{document}
\maketitle
\begin{abstract}

Commonsense knowledge acquisition is a key problem for artificial intelligence.
Conventional methods of acquiring commonsense knowledge generally require laborious and  costly human annotations, which are not feasible on a large scale.
In this paper, we explore a practical way of mining commonsense knowledge from linguistic graphs, with the goal of transferring cheap knowledge obtained with linguistic patterns into expensive commonsense knowledge.
The result is a conversion of ASER \cite{zhang2020aser}, a large-scale selectional preference knowledge resource, into TransOMCS, 
of the same representation as {\em ConceptNet} 
~\cite{liu2004conceptnet} but two orders of magnitude larger.
Experimental results demonstrate the transferability of linguistic knowledge to commonsense knowledge and the effectiveness of the proposed approach in terms of quantity, novelty, and quality.
TransOMCS is publicly available.\footnote{https://github.com/HKUST-KnowComp/TransOMCS}

\end{abstract}

\section{Introduction}

\emph{Commonsense knowledge} is defined as the
knowledge that people share but often omit when communicating
with each other.
In their seminal work, \cite{liu2004conceptnet} defined commonsense knowledge as the knowledge ``used in a technical sense to refer to the millions of basic facts and understandings possessed by most people.''
After around 20 years of development, ConceptNet 5.5~\cite{DBLP:conf/aaai/SpeerCH17}, built based on the original ConceptNet~\cite{liu2004conceptnet}, contains 21 million edges connecting
over 8 million nodes.
However, most of the knowledge assertions in ConceptNet 5.5 are still facts about entities integrated
from other knowledge sources.
The core of ConceptNet, which is inherited from the Open Mind CommonSense (OMCS) project~\cite{liu2004conceptnet}, only contains 600K pieces of high-quality commonsense knowledge in the format of tuples, e.g., (`song', \textit{UsedFor}, `sing').
The gap between the small scale of existing commonsense knowledge resources and the broad demand of downstream applications motivates us to explore richer approaches to acquire commonsense knowledge from raw text, which is cheaper and more feasible.


Throughout the history of AI, many works were developed to extract various kinds of knowledge from raw texts with human-designed linguistic patterns. For example, OpenIE~\cite{DBLP:journals/cacm/EtzioniBSW08} aims at identifying open relations between different entities (e.g., `Paris'-\textit{CapitalOf}-`France') and Hearst patterns~\cite{hearst-1992-automatic} are used to extract hyponyms (e.g., `apple'-\textit{IsA}-`fruit').
These patterns are often of high-precision. However, they typically suffer from brittleness and low coverage. 
To overcome the limitations of pattern-based methods, supervised commonsense knowledge acquisition methods were proposed.
They either model the commonsense knowledge acquisition problem as a knowledge graph completion task to predict relations between concepts~\cite{li2016commonsense} or model it as a generation task by leveraging pre-trained language models to generate tail concepts~\cite{bosselut2019comet}.
However, these approaches often are supervised with expensive annotations and 
are restricted to the distribution of the training data.

In this paper, we propose to discover new commonsense knowledge from linguistic graphs whose nodes are words and edges are linguistic relations, which is motivated by the observations~\cite{resnik1997selectional,DBLP:conf/acl/ZhangDS19} that selectional preferences over linguistic relations can reflect humans' 
commonsense about their word choice in various contexts.
Here, we use the linguistic graphs from ASER~\cite{zhang2020aser}, which is extracted from raw text with dependency parser and explicit discourse connectives and provides 27 millions of eventualities extracted using dependency patterns and 10 millions of discourse relations as its core.
Then,
we develop an algorithm for discovering patterns from the overlap of existing commonsense and linguistic knowledge bases and use a commonsense knowledge ranking model to select the 
highest-quality extracted knowledge.
As a result, we can build TransOMCS, a new commonsense knowledge graph.

In summary, our contributions are: (1) We formally define the task of mining commonsense knowledge from linguistic graphs and propose an approach to address it; (2) We construct a large-scale commonsense  resource TransOMCS, with size two orders of magnitude larger than OMCS; (3) We conduct both intrinsic and extrinsic experiments to show the value of TransOMCS.

\section{Problem Definition}

We start by defining the task of mining commonsense knowledge from linguistic graphs. Given a seed commonsense knowledge set $\CM$ (which contains $m$ tuples) and a linguistic graph set $\GM$ (which contains $n$ linguistic graphs $G$) with $m \ll n$. 
Each commonsense fact is in a tuple format ($h$, $r$, $t$) $\in \CM$,  where $r \in \RM$, the set of human-defined commonsense relations (e.g., `UsedFor', `CapableOf', `AtLocation', `MotivatedByGoal'), and $h$ and $t$ are arbitrary phrases.
Our objective is to infer a new commonsense knowledge set $\CM^+$ (with $m^+$ pieces of commonsense knowledge) from $\GM$ with the help of $\CM$ such that $m^+ \gg m$.

\vspace{-0.1in}

\section{Method}
\subsection{Overview}
\begin{figure}
    \centering
    \includegraphics[width=0.85\linewidth]{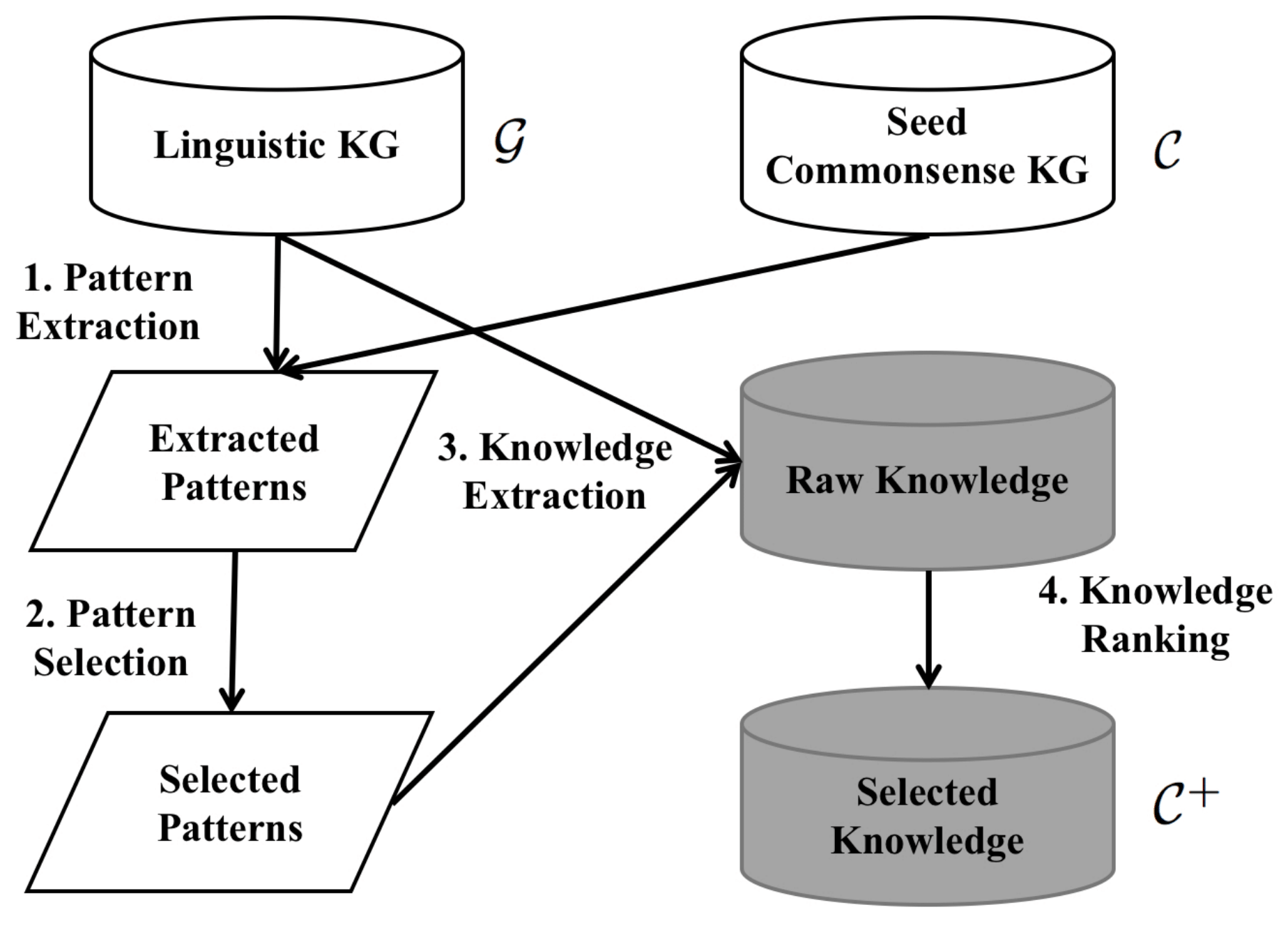}
    \vspace{-0.1in}
    \caption{The overall framework.}
    \vspace{-0.1in}
    \label{fig:framework}
\end{figure}

The proposed framework is shown in Figure~\ref{fig:framework}. In the beginning, for each seed commonsense tuple ($h$, $r$, $t$) $\in \CM$, we match and select supporting linguistic graphs that contain all the terms in $h$ and $t$, and then extract the linguistic patterns for each commonsense relation with the matched commonsense tuple and linguistic graph pairs. 
Next, we use a pattern filtering module to select the highest quality patterns. 
Finally, we train a discriminative model to evaluate the quality of the extracted commonsense knowledge. 
The details are as follows.

\vspace{-0.1in}
\subsection{Knowledge Resources}
We first introduce the details about the selected commonsense and linguistic knowledge resources. For the seed commonsense knowledge, we use the English subset of ConceptNet 5.5~\cite{DBLP:conf/aaai/SpeerCH17}. Following conventional approaches~\cite{saito2018commonsense,bosselut2019comet}, we consider only the relations covered by the original OMCS project~\cite{liu2004conceptnet}, except those with vague meanings (i.e., `RelatedTo') or those well-acquired by other knowledge resources (i.e., `IsA'\footnote{The extraction of `IsA' relations belongs to the task of hyponym detection and such knowledge has been well preserved by knowledge resources like Probase~\cite{wu2012probase}.}).
In total, 36,954 words, 149,908 concepts, and 207,407 tuples are contained in the selected dataset as $\CM$. For the linguistic knowledge resource, we use the core subset of ASER~\cite{zhang2020aser} with 37.9 million linguistic graphs\footnote{As both the internal structure of eventualities and external relations between eventualities could be converted to commonsense knowledge, we treat all the eventualities and eventuality pairs in ASER as the linguistic graphs.} to form $\GM$.

\begin{figure}[t]
    \centering
    \includegraphics[width=\linewidth]{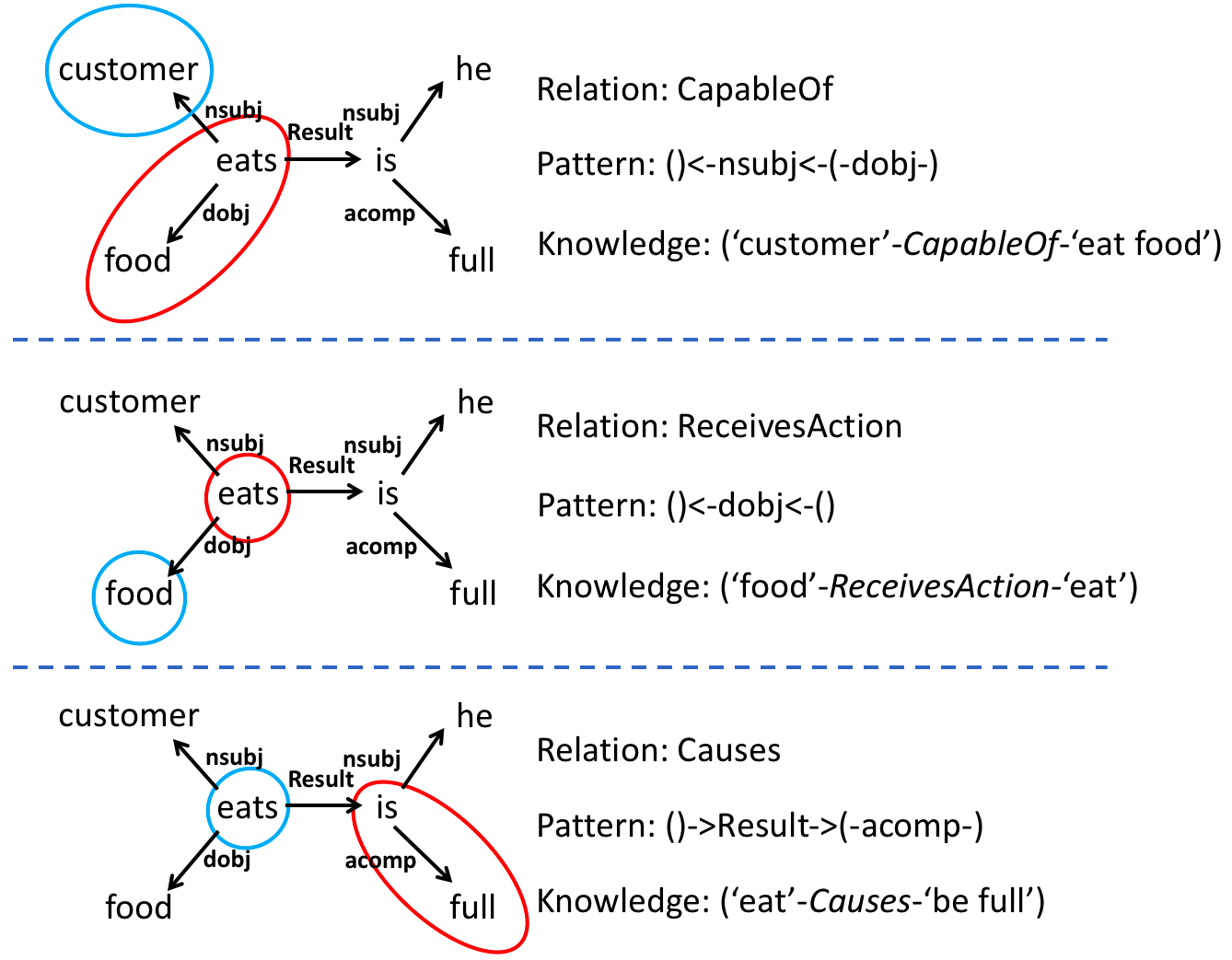}
    \vspace{-0.25in}
    \caption{Example of linguistic graphs and extracted patterns for different commonsense relations, which are extracted with the matching of words in seed commonsense tuples and the graphs.
    Given a linguistic graph as the input, these patterns can be applied to extract OMCS-like commonsense knowledge. Extracted head and tail concepts are indicated with blue and red circles respectively. 
    }
    \vspace{-0.15in}
    \label{fig:pattern_example}
\end{figure}

\subsection{Pattern Extraction}
Given a matched pair of a commonsense tuple ($h$, $r$, $t$) $\in \CM$ and a linguistic graph $G\in\GM$, the goal of the pattern extraction module is to find a pattern over linguistic relations such that given $r$, we can accurately extract all the words in $h$ and $t$ from $G$.
Here, we formally define each pattern $P$ as follows:

\begin{definition}
    Each pattern $P$ contains three components: head structure $p_h$, tail structure $p_t$, and internal structure $p_i$. $p_h$ and $p_t$ are the smallest linguistic sub-graph in $G$ that can cover all the words in $h$ and $t$, respectively. $p_i$ is shortest path from $p_h$ to $p_t$ in $G$.
\end{definition}

First, we extract $p_h$ and $p_t$ from $G$ to cover all the words in $h$ and $t$. 
We take the head pattern as an example. 
For each word in $h$, we first find its position in $G$. To avoid any ambiguity, if we find more than one match in $G$, we discard the current pair and record no pattern.
Then, with the position of the first word in $h$ as the start node, we conduct a breadth-first search (BFS) algorithm over $G$ to find a sub-structure of $G$ that covers all and only the words in $h$.
If the BFS algorithm finds such a sub-structure, we treat it as $p_h$, otherwise, we discard this example and return no pattern.
We extract the tail pattern $p_t$ with $G$ and $t$ in a similar way.
After extracting $p_h$ and $p_t$, we then collect the internal structure $p_i$, which is the shortest path from $p_h$ to $p_t$ over $G$.
To do so, we collapse all the nodes and edges in $p_h$ and $p_t$ into single `head' and `tail' nodes, respectively. Then we use `head' as the starting node to conduct a new BFS to find the shortest path from node `head' to `tail'.
We aggregate $p_h$, $p_i$, and $p_t$ together to generate the overall pattern $P$.
Examples of patterns are shown in Figure~\ref{fig:pattern_example}.
For each commonsense relation $r$, we first collect all the commonsense tuples of relation $r$ in $\CM$ to form the subset $\CM^r$. 
Then for each $(h,r,t) \in \CM^r$, we go over the whole $\GM$ to extract matched patterns with the aforementioned algorithm. The time complexity of the overall algorithm is O($|\CM| \cdot |\GM| \cdot N^2$), where $|\CM|$ is the size of $\CM$, $|\GM|$ is the size of $\GM$ and $N$ is the maximum number of the nodes in $G$.

\subsection{Pattern Selection and Knowledge Extraction}
We evaluate the plausibility of pattern $P$ regarding commonsense relation $r$ as follows:
\begin{equation}
    P(P|r) = \frac{F(P|r)}{\sum_{P^\prime \in \PM^r}F(P^\prime|r)},
\end{equation}
where $F(P|r)$ is the scoring function we use to determine the quality of $P$ regarding  $r$ and $\PM^r$ is the set of patterns extracted for $r$.
We design $F(P|r)$ as follows:
\begin{equation}
    F(P|r) = C(P|r) \cdot L(P) \cdot U(P|r),
\end{equation}
where $C(P|r)$ indicates counts of observing $P$ regarding $r$, $L(P)$ indicates the length of $P$ to encourage complex patterns, and $U(P|r) = \frac{C(P|r)/\sqrt{|\CM^r|}}{\sum_{r^\prime \in \RM}C(P|r^\prime)/\sqrt{|\CM^{r^\prime}|}}$ is the uniqueness score of $P$ regarding $r$.
We select the patterns with the plausibility score higher than 0.05.
On average, 2.8 high confident patterns are extracted for each relation.
After extracting the matched patterns, we then apply the extracted patterns to the whole linguistic graph set $\GM$ to extract commonsense knowledge.
For each $G \in \GM$ and each relation $r$, we go through all the extracted patterns $P \in \PM^r$. If we can find a matched $P$ in $G$, we will then extract the corresponding words of $p_h$ and $p_t$ in $G$ as the head words and tail words, respectively. The extracted head and tail word pairs are then considered as a candidate knowledge, related via $r$.




\subsection{Commonsense Knowledge Ranking}

To minimize the influence of the pattern noise, we propose a knowledge ranking module to rank all extracted knowledge based on the confidence.
To do so, we first use human annotators to label a tiny subset of extracted knowledge and then use it to train a classifier. We assign a confidence score to each extracted knowledge via the learned classifier. 

\subsubsection{Dataset Preparation}
For each commonsense relation type, we randomly select 1,000 tuples\footnote{For relation `DefinedAs' and `LocatedNear', as our model only extracted 26 and 7 tuples respectively, we annotate all of them and exclude them when we compute the overall accuracy in Section~\ref{sec:intrinsic}.} to annotate.
For each tuple, five annotators from Amazon Mechanical Turk are asked to decide if they think the tuple is a plausible commonsense statement.
If at least four annotators vote for plausibility, we label that tuple as a positive example.
Otherwise, we label it as a negative example.
Additionally, we include all the matched examples from OMCS as positive examples.
In total, we obtain 25,923 annotations for 20 commonsense relations. 
Among these annotations, 18,221 are positive examples and 7,702 are negative examples. 
On average, each tuple has 12.7 supporting linguistic graphs.
We randomly select 10\% of the dataset as the test set and the rest as the training set.\footnote{As the annotated dataset is slightly imbalanced, when we randomly select the test set, we make sure the number of positive and negative examples are equal.}

\begin{figure}[t]
    \centering
    \includegraphics[width=0.8\linewidth]{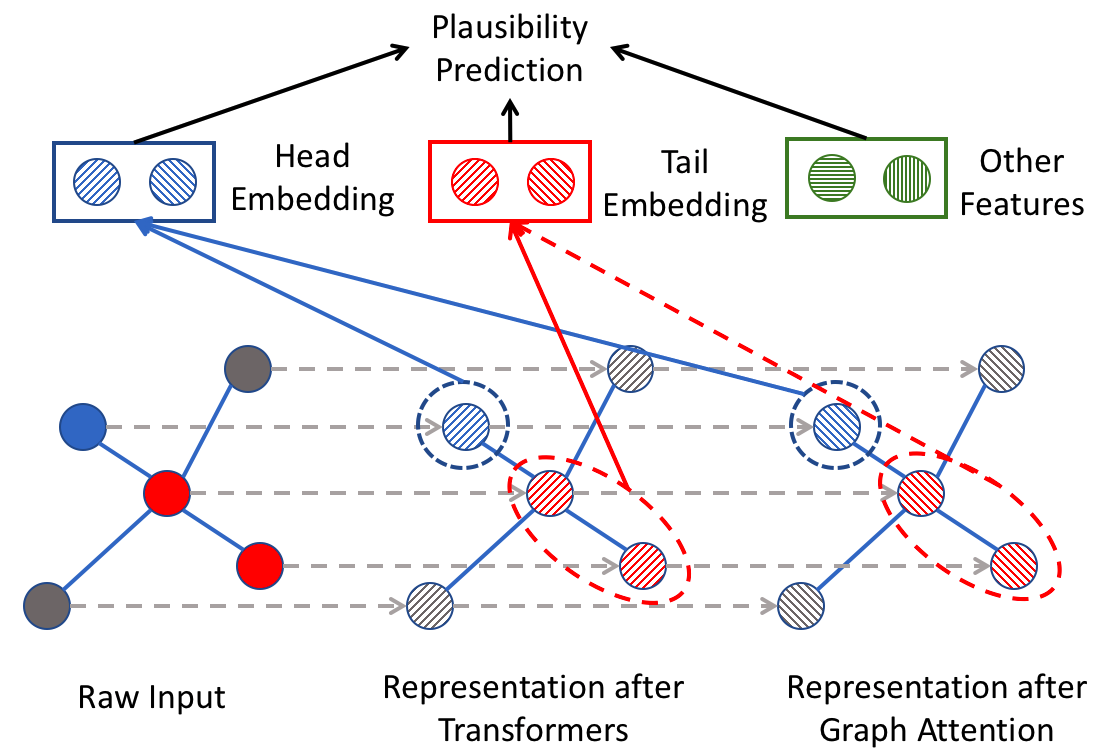}
    \vspace{-0.1in}
    \caption{Demonstration of the model. The blue and red colors denote the head words and tail words, respectively. The gray and green colors and indicate the other words and features, respectively.}
    \label{fig:classification_model}
    \vspace{-0.1in}
\end{figure}

\subsubsection{Task Definition} 
The goal of this module is to assign confidence scores to all extracted knowledge so that we can rank all the knowledge based on their quality.
As for each extracted tuple, we may observe multiple supporting linguistic graphs.
We model it as a multi-instance learning problem. Formally, we define the overall annotated knowledge set as $\KM$. For each $k \in \KM$, denote its supporting linguistic graph set as $\GM^k$. 
We use $F(k|\GM^k)$ to denote the plausibility scoring function of $k$ given $\GM^k$, and it can be defined as follows:
\begin{equation}
    F(k|\GM^k) = \frac{1}{|\GM^k|} \cdot \sum_{g\in\GM^k}f(k|g),
\end{equation}
where $|\GM^k|$ is the number of graphs in $\GM^k$ and $f(k|g)$ is the plausibility score of $k$ given $g$.

\subsubsection{Model Details}

The proposed model, as shown in Figure~\ref{fig:classification_model}, contains three components: transformer, graph attention and the final plausibility prediction.

\noindent \textbf{Transformer:} 
We use the transformer module as the basic layer of our model. Formally, assuming $g$ contains $n$ words $w_1, w_2, ..., w_n$, we denote the representation of all words after the transformer module\footnote{For technical details of the Transformer network, you may refer to the original paper~\cite{vaswani2017attention}.} as $\e_1, \e_2, ..., \e_n$. In our model, we adopt the architecture of BERT~\cite{DBLP:conf/naacl/DevlinCLT19} and their pre-trained parameters (BERT-base) as the initialization.

\noindent \textbf{Graph Attention:} Different from conventional text classification tasks, linguistic relations play a critical role in our task.
Thus in our model, we adopt the graph attention module~\cite{DBLP:conf/iclr/VelickovicCCRLB18} to encode the information of these linguistic relations.
For each $w$ in $g$, we denote its representation as $\hat{\e}$, which is defined as follows:
\begin{equation}
    \hat{\e} = \sum_{\e^\prime \in N(\e)} a_{\e, \e^\prime} \cdot \e^\prime,
\end{equation}
where $N(e)$ is the representation set of words that are connected to $w$ and $a_{\e, \e^\prime}$ is the attention weight of $\e^\prime$ regarding $\e$.
Here, we define the attention weight as:
\begin{equation}
    a_{\e, \e^\prime} = \frac{e^{\text{NN}_a([\e, \e^\prime])}}{\sum_{\bar{\e} \in N(\e)} e^{\text{NN}_a([\e, \bar{\e}])}},
\end{equation}
where $[.]$ indicates vector concatenation and $\text{NN}_a$ is the dense neural network we use to predict the attention weight before the softmax module.
After the graph attention module, 
the representation of words are then denoted as $\hat{\e_1}, \hat{\e_2}, ..., \hat{\e_n}$.

\noindent \textbf{Plausibility Prediction:}
In the last part of our model, we first concatenate $\e$ and $\hat{\e}$ together for all the words and then create head embedding $\o_{head}$ and tail embedding $\o_{tail}$ using mean pooling over $[\e, \hat{\e}]$ of all words appear in the head or tail respectively. Besides embedding features, two important features, graph frequency $\o_{fre}$ (how many times this graph appears) and graph type $\o_{type}$ (whether this graph is node or edge), are also considered for the final plausibility prediction. 
We define plausibility prediction function $f(k|g)$ as:
\begin{equation}
    f(k|g) = \text{NN}_p([\o_{head}, \o_{tail}, \o_{fre}, \o_{type}]),
\end{equation}
where $NN_p$ is a fully connected layer we use to predict the plausibility.
We use the cross-entropy as the loss function and stochastic gradient descent as the optimization method.

\subsubsection{Model Performance} We train the classifier for different relations separately as some relations are not exclusive with each other (e.g., `RecievesAction' and `UsedFor') and test our model on our collected data. We compare our model with random guess and directly using BERT, which is identical to our model excluding the graph attention part, in Table~\ref{tab:model_performance}. 
Experimental results show that both the Transformer and the graph attention modules make significant contributions to the prediction quality of the extracted knowledge.
\begin{table}[h]
    \centering
    \small
    \begin{tabular}{c||cc|c}
    \toprule
         & Random & BERT & Proposed model \\
        \midrule
        Accuracy & 50\% & 70.91\% & 73.23\% \\
    \bottomrule
    \end{tabular}
    \caption{Performance of different plausibility prediction models.}
    \label{tab:model_performance}
    \vspace{-0.2in}
\end{table}

\section{Intrinsic Evaluation}\label{sec:intrinsic}
We compare our approach with recently developed commonsense knowledge acquisition methods COMET~\cite{bosselut2019comet} and LAMA~\cite{petroni2019language}.


\begin{table*}[t]
\small
    \centering
    \resizebox{0.72\textwidth}{!}{%
    \begin{tabular}{l||cc|cc|cc}
    \toprule
    Model & \# Vocab & \# Tuple & Novel$_t$ & Novel$_c$ & ACC$_n$ & ACC$_o$\\
    \midrule
        COMET$_{Original}$ (Greedy decoding)        & 715   & 1,200   & 33.96\% & 5.27\%  & 58\% & 90\% \\
        COMET$_{Original}$ (Beam search - 10 beams) & 2,232 & 12,000  & 64.95\% & 27.15\% & 35\% & 44\% \\
    \midrule
        COMET$_{Extended}$ (Greedy decoding)        & 3,912 & 24,000  & \textbf{99.98\%} & 55.56\% & 34\% & 47\% \\
        COMET$_{Extended}$ (Beam search - 10 beams) & 8,108 & 240,000 & \textbf{99.98\%} & 78.59\% & 23\% & 27\% \\
        \midrule
        LAMA$_{Original}$ (Top 1)        &  328  & 1,200   & - &  - & - & 49\% \\
        LAMA$_{Original}$ (Top 10)       &  1,649  & 12,000   &-  & -  & - & 20\% \\
    \midrule
        LAMA$_{Extended}$ (Top 1)      & 1,443 & 24,000  & - & - & - & 29\% \\
        LAMA$_{Extended}$ (Top 10)     & 5,465 & 240,000  & - & - & - & 10\% \\
    \midrule
    TransOMCS$_{Original}$ (no ranking)       & 33,238 & 533,449 & 99.53\% & 89.20\% & 72\% & 74\%\\
    \midrule
        TransOMCS (Top 1\%)         & 37,517 & 184,816 & 95.71\% & 75.65\% & \textbf{86\%} & 87\% \\
        TransOMCS (Top 10\%)        & 56,411 & 1,848,160 & 99.55\% & 92.17\% & 69\% & 74\% \\
        TransOMCS (Top 30\%)        & 68,438 & 5,544,482 & 99.83\% & 95.22\% & 67\% & 69\% \\
        TransOMCS (Top 50\%)        & 83,823 & 9,240,803 & 99.89\% & 96.32\% & 60\% & 62\% \\
    \midrule
    TransOMCS (no ranking)     & \textbf{100,659} & \textbf{18,481,607} & 99.94\% & \textbf{98.30\%} & 54\% & 56\%\\
    \midrule
       OMCS in ConceptNet 5.0                    & 36,954 & 207,427   & - & - & - & \textbf{92\%} \\
    \bottomrule
    \end{tabular}
    }
    \caption{Main Results. For our proposed method, we present both the full TransOMCS and several subsets based on the plausibility ranking scores. TransOMCS$_{Original}$ means the subset, whose head/relation appear in the test set as used by COMET$_{Original}$ and LAMA$_{Original}$.  As LAMA is unsupervised, the Novelty metric is not applicable. For our model, for the fair comparison, we exclude all annotated knowledge.}
    \label{tab:main_result}
    \vspace{-0.15in}
\end{table*}

\subsection{Evaluation Metrics}

\noindent \textbf{Quantity:} We first measure \emph{quantity} of the algorithms.
Specifically, we are interested in two metrics: the number of acquired commonsense knowledge tuples and the number of unique words. 

\noindent \textbf{Novelty:} 
For measuring \emph{novelty} of the outputs, we follow~\cite{bosselut2019comet} and report the proportion of novel tuples and concepts, denoted as Novel$_t$ and Novel$_c$, respectively. 
Here `novel', similar to the definition in~\cite{bosselut2019comet}, means that one cannot find the whole tuple or concept in the training/development data via string match. 
For COMET and LAMA, as the head concept is given as input and thus only the tail concept is generated by the models, we only select the tail concepts to calculate the concept novelty.

\noindent \textbf{Quality:} We use Amazon Mechanical Turk to evaluate the \emph{quality} of the acquired commonsense tuples.  
For each model, we randomly select 100 tuples from the overall set to test the quality.
For each tuple, five workers are invited to annotate whether this tuple is plausible or not. If at least four annotators label the tuple as plausible, we then consider it as a plausible tuple.
Additionally, to investigate the quality of tuples with novel concepts, for each model, we also report the performance of all sampled novel tuples.
As we use the accuracy (\# valid tuples/ \# all tuples) to evaluate the quality, we denote the overall quality of tuples with novel concepts and the whole tuple set as ACC$_n$ and ACC$_o$, respectively.

\subsection{Baseline Methods}
\begin{enumerate}[leftmargin=*]
\setlength{\itemsep}{0pt}
\setlength{\parsep}{0pt}
\setlength{\parskip}{0pt}
    \item \textbf{COMET:} Proposed by \cite{bosselut2019comet}, COMET leverages the pre-trained language model GPT~\cite{radford2018improving} to learn from annotated resources to generate commonsense knowledge.
    \item \textbf{LAMA:} Proposed by \cite{petroni2019language}, LAMA leverages  BERT~\cite{DBLP:conf/naacl/DevlinCLT19} to acquire commonsense knowledge. 
    Unlike COMET, LAMA is unsupervised.
\end{enumerate}


We denote our method as \textbf{TransOMCS}, indicating that we transfer knowledge from linguistic patterns to OMCS-style commonsense knowledge.


\subsection{Implementation Details}


For both COMET and LAMA, 
we include two variants in our experiments. 
The first one follows COMET's paper and uses concept/relation pairs among the most confident subset of OMCS as input. 
Due to the small size of this subset (only 1.2K positive examples), even if we use the 10-beam search decoding, we can only generate 12K tuples. 
To overcome this limitation and test whether these models can be used for generating large-scale commonsense knowledge, we consider a different evaluation setting where we randomly select 24K  concepts from the concepts extracted by our approach and then randomly pair the selected concepts with commonsense relations as the input.
We refer to the first and modified settings with $_{Original}$ and $_{Extended}$ subscripts, respectively.


\subsection{Result Analysis}

The summary of model evaluations is listed in Table~\ref{tab:main_result}. 
Compared with all baseline methods in their original settings, TransOMCS outperforms them in quantity. Even the smallest subset (top 1\%) of TransOMCS outperforms their largest generation strategy (10-beam search) by ten times. 
TransOMCS also outperforms COMET in terms of novelty, especially the percentage of novel concepts. The reason behind this is that COMET is a pure machine-learning approach and it learns to generate the tail concept in the training set. It is possible that the stronger their models are, the more likely they overfit the training data, the fewer novel concepts are generated. 
As for the quality, when the training data is similar to the test data, COMET provides the best quality. For example, in the original setting, the greedy decoding strategy achieves 90\% overall accuracy. As a comparison, the quality of LAMA is less satisfying, which is mainly because LAMA is fully unsupervised. Besides that, in the extended setting, the quality of both models drops, which is mainly because the randomly generated pairs could include more rare words. Compared with them, TransOMCS (top 1\%) provides the comparable quality with COMET, but with a larger scale. 
When the quantity is comparable, TransOMCS outperforms both of them in terms of quality.
We summarize the comparisons in Table~\ref{tab:comparison}.

\begin{table}[t]
	\small
	\centering
	\resizebox{0.85\linewidth}{!}{%
	\begin{tabular}{c||c|c|c|c}
		\toprule
		& Annotation & Scale &  Novelty & Quality \\
		\midrule
		OMCS     & full & small & high & high \\
		\midrule
		COMET & supervision  & large & low & depends\\
		LAMA & no  & large & high & low \\
		\midrule
		TransOMCS & supervision & large  & high & high \\
		\bottomrule
	\end{tabular}
	}
	\caption{Comparison of different commonsense resources or acquisition methods. The quality of COMET depends on the scale of its generated knowledge (high quality on the test set, but low quality on the large scale generation.) COMET and TransOMCS require a small number of annotations as supervision.}
	 \vspace{-0.2in}
	\label{tab:comparison}
\end{table}

\subsection{Case Study}
\begin{figure*}[tb]
	\centering
	\includegraphics[width=0.93\linewidth]{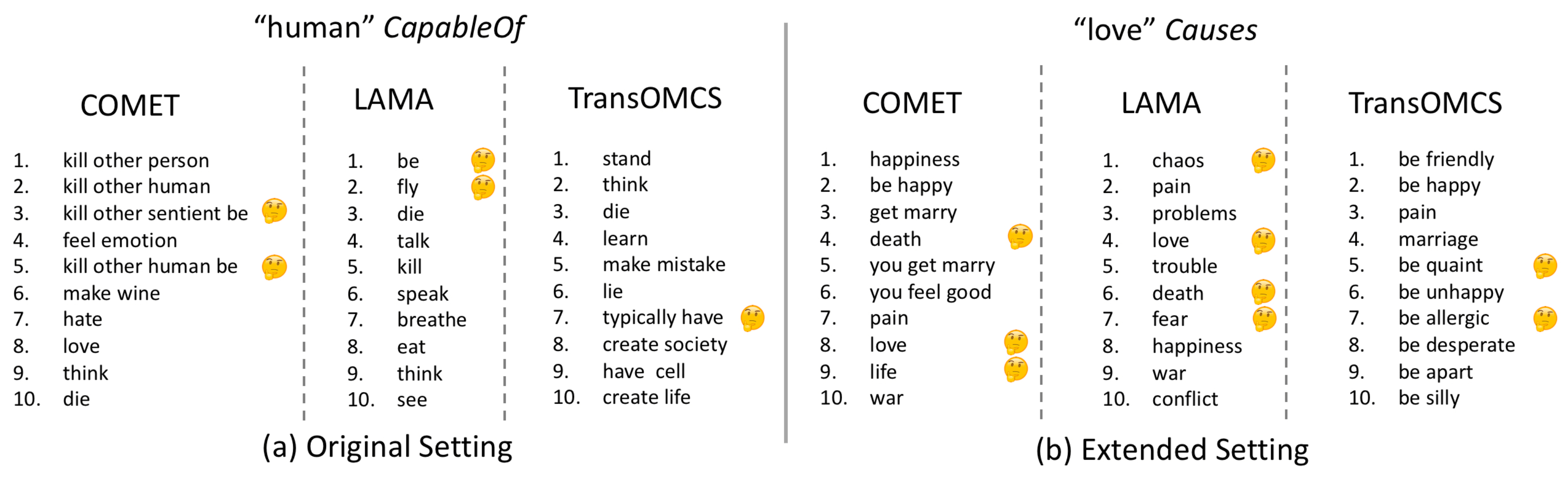}      		  		
	\vspace{-0.1in}
	\caption{Case study of three knowledge acquisition methods. Two head-relation pairs
	are selected from the original and extended settings respectively. We indicate low quality tuples with confusing face emojis. For each model, we select the top ten most confident results.	} 
	\vspace{-0.05in}
	\label{fig:case_study}
\end{figure*}

We show case studies in Figure~\ref{fig:case_study} to further analyze different acquisition methods.

\begin{table*}[t]
	\centering
	\small
	\resizebox{0.68\linewidth}{!}{%
	\begin{tabular}{l||c|c||c|c}
		\toprule
		\multirow{2}{*}{Commonsense Knowledge Resource} & \multicolumn{2}{c||}{Reading Comprehension} & \multicolumn{2}{c}{Dialog Generation}\\
		\cline{2-5}
		& Accuracy (\%) & $\Delta$ (\%) & BLEU & $\Delta$\\
		\midrule
		Base model (no external knowledge resource)         & 82.90 & -     & 0.54 & - \\
		\midrule
		\quad +OMCS                                   & 83.11 & +0.21 & 0.72 & +0.18 \\
		\midrule
		\qquad+COMET$_{Original}$ (Greedy decoding)          & 83.12 & +0.22 & 0.61 & +0.07 \\
		\qquad+COMET$_{Extended}$ (Beam search - 10 beams)  & 83.03 & +0.13 & 0.68 & +0.14 \\
		\midrule
		\qquad+LAMA$_{Original}$ (Top 1)                     & 83.13 & +0.23 & 0.56 & +0.02 \\
		\qquad+LAMA$_{Extended}$ (Top 10)                   & 83.17 & +0.27 & 0.57 & +0.03 \\
		\midrule
		\qquad+TransOMCS (Top 1\%)                    & \textbf{83.27} & \textbf{+0.37} & \textbf{1.85} & \textbf{+1.31}\\
		\bottomrule
	\end{tabular}
	}
\vspace{-0.05in}
	\caption{Experimental results on downstream tasks. For COMET and LAMA, we report the performance of their most accurate and largest setting. For TransOMCS, as current models cannot handle large-scale data, we only report the performance of the most confident 1\%. All the numbers are computed based on the average of four different random seeds rather than the best seed as reported in the original paper. }
	\vspace{-0.15in}
	\label{tab:downstream}
\end{table*}

\noindent \textbf{COMET:} COMET is the only one that can generate long concepts. At the same time, it also suffers from generating meaningless words. For example, two of the generated concepts end with `be', which is confusing and meaningless. This is probably because COMET only generates lemmas rather than normal words. Besides that, COMET could overfit the training data, even though the ten outputs are not exactly the same, four of them mean the same thing (`kill others').

\noindent \textbf{LAMA:} The most significant advantage of LAMA is that it is unsupervised. However, it has two major drawbacks: (1) it can only generate one-token concepts, which is far away from enough for commonsense knowledge; (2) the quality of LAMA is not as good as the other two methods.

\noindent \textbf{TransOMCS:} Compared with COMET, TransOMCS can generate more novel commonsense knowledge. For example, our model knows that `human' is capable of having cells and creating life. Besides that, unlike LAMA, TransOMCS can generate multi-token concepts. At the same time, our approach also has two limitations: (1) it cannot extract long concepts, which are difficult to find an exact pattern match; (2) as the extraction process strictly follows the pattern matching, it could extract semantic incomplete knowledge. For example, `human' is capable of `have'. The original linguistic graph should be ``human have something'', but as the pattern is `()$<$-nsubj$<$-()', the object is missing.

\section{Extrinsic Evaluation}

We conduct experiments on two downstream tasks: \emph{commonsense reading-comprehension}~\cite{ostermann2018semeval} and \emph{dialogue generation}~\cite{li2017dailydialog}.
We select \cite{wang2018yuanfudao} and \cite{DBLP:conf/emnlp/LuongPM15,zhang2020aser} as our base models for the comprehension and the generation task, respectively.
For the fair comparison,
we keep all the model architecture and parameters the same
across different trials.
We 
report the results with the common metric of each dataset:
\emph{accuracy} on the comprehension task and the \emph{BLEU} score~\cite{papineni2002bleu} on the dialog generation task.

The overall result is shown in Table~\ref{tab:downstream}. 
For the reading comprehension task, 
adding the top 1\% of TransOMCS contributes  0.37 overall accuracy, compared to 0.21 contribution of OMCS. 
Meanwhile, the contributions of COMET and LAMA are minor for this task.
For the dialogue generation task,
TransOMCS shows remarkable improvement in
the quality of generated responses.
At the same time, adding other knowledge resources to OMCS does not provide any meaningful improvements to the performance. 
The reason behind this
could be
that COMET and LAMA 
provide limited 
high quality novel commonsense knowledge.
For example, the original OMCS on average contributes 1.46 supporting tuples\footnote{Here by supporting tuple, we mean that the head and tail concept appear in the post and response respectively.} and TransOMCS contribute another 3.36 supporting tuples. As a comparison, COMET$_{Original}$, COMET$_{Extended}$, LAMA$_{Original}$, and LAMA$_{Extended}$ only provide 0.01, 0.07, 0.49, 0.01 additional tuples respectively.
This experiment result shows that TransOMCS has more novel knowledge.



\section{Related Work}

Commonsense knowledge covers a variety of knowledge types like knowledge about 
typical location or causes of events in OMCS~\cite{liu2004conceptnet}, 
events causes, effects, and temporal properties in ATOMIC~\cite{sap2019atomic} and ~\cite{ben2020temporal}, and physical attributes of objects ~\cite{DBLP:conf/acl/ElazarMRBR19}.
As an important knowledge resource for AI systems, commonsense knowledge has been shown helpful in many downstream tasks such as question answering~\cite{lin2019kagnet} and reading comprehension~\cite{wang2018yuanfudao}. 
Conventional commonsense resources (e.g., OMCS and ATOMIC) are often constructed via crowdsourcing
aimed to provide high-quality results; although such expensive processes restrict their scale. 
Recently, several attempts~\cite{li2016commonsense,DBLP:conf/emnlp/DavisonFR19} have been made to enrich the existing commonsense knowledge by learning to predict new relations between concepts.
However, these approaches cannot generate new nodes (concepts).
To address this problem, several models were proposed to generate commonsense tuples in either supervised ~\cite{bosselut2019comet} or unsupervised ~\cite{petroni2019language} fashions.
Different from them, 
this work shows 
that linguistic patterns can be extracted for commonsense
relations and consequently be used for producing valuable commonsense knowledge. 

\section{Conclusion}
In this paper, we exhibit
the transferability from linguistic knowledge (dependency and discourse knowledge) to commonsense knowledge (i.e., the OMCS-style knowledge) by showing that a large amount of high-quality commonsense knowledge can be extracted from linguistic graphs. 
We formally define the task of mining commonsense knowledge from linguistic graphs
and
present TransOMCS, a commonsense knowledge resource  
extracted from linguistic graphs into the format of 
the OMCS subset of ConceptNet, but two orders of magnitude larger than the original OMCS.
While TransOMCS is noisier than OMCS, it can still make significant contributions to downstream tasks due to its larger coverage, as evident by the extrinsic experiments.

\section*{Acknowledgements}

This paper was supported by the Early Career Scheme (ECS, No. 26206717) from the Research Grants Council in HK and the Tencent AI Lab Rhino-Bird Focused Research Program, with partial support from DARPA grant FA8750-19-2-1004. 

\clearpage

\bibliographystyle{named}
\bibliography{ref}

\end{document}